\pdfoutput=1

\documentclass[11pt]{article}

\usepackage[final]{acl}
\usepackage{amsmath} 

\usepackage{graphicx}
\usepackage{multirow}

\usepackage{url}
\usepackage{booktabs}
\usepackage{amssymb}
\usepackage{bbding}
\usepackage{pifont}
\usepackage{wasysym}
\usepackage{times}
\usepackage{latexsym}

\usepackage[T1]{fontenc}

\usepackage[utf8]{inputenc}

\usepackage{microtype}

\usepackage{inconsolata}

\usepackage[normalem]{ulem}
\usepackage{algorithm}
\usepackage{multirow}
\usepackage{enumitem}

\usepackage{booktabs}

\usepackage{algorithm}
\usepackage{algpseudocode}
\usepackage[most]{tcolorbox}

\title{Benchmarking LLM-as-a-Judge for Long-Form Output Evaluation}

\usepackage{authblk}

\author[1,2]{\textbf{Junjie Chen}\thanks{chenjj826@gmail.com}}
\author[3]{\textbf{Yuxi Dong}}
\author[1]{\textbf{Haitao Li}}
\author[1]{\textbf{Weihang Su}}
\author[1]{\textbf{Yujia Zhou}}
\author[1]{\\ \textbf{Min Zhang}}
\author[1]{\textbf{Yiqun Liu}}
\author[2,1]{\textbf{Qingyao Ai}\thanks{Corresponding Author: aiqy@tsinghua.edu.cn}}

\affil[1]{Department of Computer Science and Technology, Tsinghua University}
\affil[2]{Quancheng Laboratory}
\affil[3]{University of Science and Technology Beijing}

\begin{document}
\maketitle

\begin{abstract}
As large language models (LLMs) are increasingly used for long-form generation, reliably evaluating long-form outputs has become a critical challenge. 
LLM-as-a-judge offers a scalable alternative to human evaluation, yet its reliability in long-form output evaluation remains underexamined: existing meta-evaluation benchmarks focus mainly on short-form outputs. Compared with short-form evaluation, long-form evaluation is not merely a matter of output length; it often requires judges to handle more complex document-level demands.
In this work, we introduce \textit{LongJudgeBench}, a comprehensive benchmark for evaluating LLM judges on long-form outputs across diverse real-world scenarios and judging protocols.
We systematically evaluate a broad range of LLM judges, covering multiple base models and judging settings. 
Our results reveal a substantial reliability gap: current LLM judges remain unstable across scenarios, and rubrics or references are helpful but not always sufficient. 
We hope \textit{LongJudgeBench} will support future research on more robust, context-aware, and human-aligned LLM-as-a-judge methods. 
Our code is available at \url{https://github.com/cjj826/LongJudgeBench}.
\end{abstract}

\section{Introduction}
As large language models (LLMs)~\cite{chang2024survey,guo2025deepseek} continue to advance in open-ended generation tasks such as deep research and creative writing, reliable and efficient evaluation of their outputs has become an increasingly critical challenge. Existing evaluation methods generally fall into manual and automatic approaches. Manual evaluation~\cite{zheng2023judging} is often considered reliable, but also costly, time-consuming, and difficult to scale with the rapid pace of model development. Automatic evaluation~\cite{chang2024survey} offers better scalability, yet conventional metrics remain limited in assessing open-ended responses, whose quality is inherently multidimensional. These limitations have motivated the growing adoption of LLM-as-a-judge~\cite{li2024llms}, a flexible and cost-effective paradigm in which LLMs assess the quality of candidate outputs.

\begin{table}[t]
\resizebox{\columnwidth}{!}{%
\begin{tabular}{lrrr}
\hline
Benchmark                                 & \# Inst. & Mean  & Med. \\ \hline
MT-Bench~\cite{zheng2023judging}          & 4,796   & 194.0 & 142  \\
LLMBar~\cite{zeng2024evaluating}          & 419     & 106.0 & 49   \\
RewardBench~\cite{lambert2025rewardbench} & 8,108   & 194.8 & 115  \\
PreferenceBench~\cite{kim2024prometheus}  & 1,998   & 177.2 & 168  \\
RPR~\cite{pitis2024improving}             & 11,167  & 94.0  & 89   \\
JudgeBench~\cite{tan2025judgebench}       & 620     & 446.0 & 387  \\
RewardBench 2~\cite{malik2026rewardbench}  & 1,865   & 312.3 & 265  \\ \hline
\end{tabular}%
}
\caption{Statistics of candidate output lengths in existing LLM-as-a-judge benchmarks. \# Inst., Mean, and Med. denote instance count, mean length, and median length, respectively. Length is measured in tokens using the \texttt{cl100k\_base} tokenizer from \texttt{tiktoken}.}
\label{tab:response_length_stats}
\end{table}

As LLM-as-a-judge becomes increasingly central to model evaluation, assessing the reliability of the judges themselves has emerged as an important research problem. 
To this end, prior work has introduced a range of meta-evaluation benchmarks that compare model judgments against human preferences or other high-quality annotations~\cite{lambert2025rewardbench,zheng2023judging,kim2024prometheus,malik2026rewardbench,tan2025judgebench,zeng2024evaluating,pitis2024improving}. 
These benchmarks have provided valuable evidence on the promise and limitations of LLM judges, exposing recurring issues such as position bias, length bias, and unstable judgments on complex tasks. 

However, the growing use of LLMs for long-form tasks, such as deep research, report generation, and content creation, reveals an important gap in existing meta-evaluation settings. 
As shown in Table~\ref{tab:response_length_stats}, most prior benchmarks are built around candidate outputs of only a few hundred tokens, leaving limited evidence about judge reliability in long-form evaluation. 
This gap is not merely a matter of output length: long-form evaluation involves qualitatively different document-level evaluation demands, such as assessing overall organization, cross-section consistency, and coverage and depth across multiple subtopics.
This raises a central question: \textit{Can LLM-as-a-judge reliably evaluate long-form outputs in real-world tasks?}

To answer this question, we introduce \textit{LongJudgeBench}, a comprehensive meta-evaluation benchmark for assessing LLM-as-a-judge in long-form output evaluation. Unlike existing judge benchmarks that mainly focus on short-form responses, \textit{LongJudgeBench} targets long-form outputs across five representative real-world scenarios and six datasets. It provides candidate outputs averaging 9,249.7 tokens, together with task-specific evaluation protocols and expert reference judgments. 
\textit{LongJudgeBench} is highlighted in the following two aspects:

\textbf{(1) Real-world scenarios for long-form output evaluation.}
\textit{LongJudgeBench} covers five practical scenarios, including deep research, scientific survey, creative writing, long-chain analysis, and systematic review. These scenarios involve diverse document-level evaluation demands, such as research coverage, survey structure and content quality, writing preferences, and domain-specific insight assessment, making the benchmark closer to real-world long-form evaluation needs.

\textbf{(2) Diverse tasks with reliable expert judgments.}
\textit{LongJudgeBench} integrates six datasets with diverse output lengths, difficulty levels, and judgment protocols. For reliable meta-evaluation, all reference judgments come from carefully collected expert annotations, including direct scoring, preference labeling, ranking, and reference verification, with necessary quality control.

Using \textit{LongJudgeBench}, we evaluate a variety of LLM judges across different model families and judging settings. Our experiments reveal a clear reliability gap in current LLM-as-a-judge methods under long-form evaluation, highlighting the need for more robust, context-aware, and human-aligned LLM-as-a-judge methods.

\begin{figure*}[t]
    \centering
    \includegraphics[width=\textwidth]{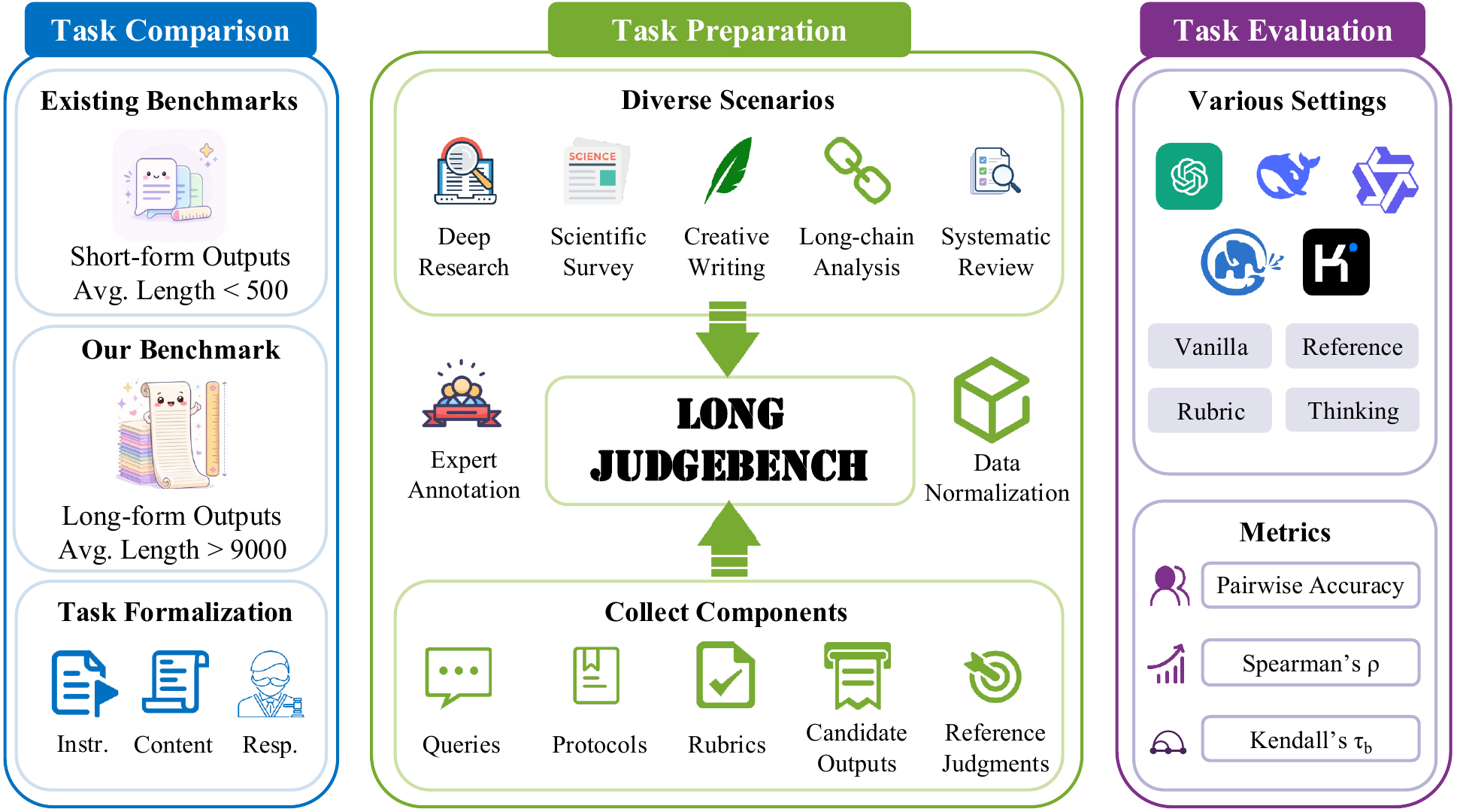}
    \caption{Overview of \textit{LongJudgeBench}. Unlike existing LLM-as-a-judge benchmarks focused on short-form outputs, \textit{LongJudgeBench} targets long-form output evaluation. It formalizes each instance with instruction, content, and response, constructs tasks from diverse scenarios with queries, protocols, rubrics, candidate outputs, and reference judgments, and evaluates different judging settings using various agreement metrics.}
    \label{fig:1}
\end{figure*}

\section{Related Work}

\textbf{LLM-as-a-judge.}
Evaluation methods for LLMs can be broadly categorized into manual and automatic approaches. Manual evaluation, such as Chatbot Arena~\cite{zheng2023judging}, is often reliable but costly and difficult to scale. Automatic evaluation includes reference-based metrics such as BLEU~\cite{papineni2002bleu}, ROUGE~\cite{lin2004rouge}, and BERTScore~\cite{zhang2019bertscore}, which rely on reference answers and capture only limited aspects of response quality, as well as multiple-choice evaluation, which is easy to score but poorly suited to open-ended tasks. LLM-based evaluation offers a more flexible alternative by using LLMs as judges to score, compare, or critique candidate outputs, and its promising agreement with human preferences in some open-ended settings has led to broad adoption in model evaluation~\cite{chu2024automatic,chen2026auto}.

\textbf{Benchmarks for LLM-as-a-judge.}
As LLM-as-a-judge has become widely used for evaluation, recent work has begun to evaluate the judges themselves. MT-Bench~\cite{zheng2023judging} examines judge--human agreement in multi-turn question answering, while LLMBar~\cite{zeng2024evaluating} tests robustness to adversarial instruction-following errors. RewardBench~\cite{lambert2025rewardbench} and RewardBench 2~\cite{malik2026rewardbench} evaluate reward models and LLM judges using prompt--chosen--rejected triples, Preference Bench~\cite{kim2024prometheus} studies criteria-aware pairwise judgments with references and rubrics, RPR~\cite{pitis2024improving} focuses on context-dependent preference shifts, and JudgeBench~\cite{tan2025judgebench} evaluates objective correctness in knowledge, reasoning, mathematics, and code. Together, these benchmarks cover preference alignment, instruction following, reward modeling, context sensitivity, bias robustness, and objective correctness. However, they mainly evaluate short candidate outputs (Table~\ref{tab:response_length_stats}), leaving the reliability of LLM-as-a-judge for long-form output evaluation largely underexplored. Our work addresses this gap. A concurrent work, Long-form RewardBench~\cite{huang2026long}, evaluates reward models for long-form generation using chosen--rejected preference data. However, it is less suitable for benchmarking LLM-as-a-judge methods themselves, since most of its gold labels are produced by an LLM-as-a-judge pipeline rather than expert annotation, potentially causing circularity and judge-specific biases. Moreover, its responses average around 3K tokens, far shorter than the 9K+ average outputs in our \textit{LongJudgeBench}.

\section{LongJudgeBench}

\subsection{Overview}
The central question of this work is whether LLM-as-a-judge can reliably evaluate long-form outputs. To answer this question, we introduce \textit{LongJudgeBench} (as shown in Figure~\ref{fig:1}), a meta-evaluation benchmark for studying LLM-as-a-judge across diverse long-form evaluation tasks. 

We organize each evaluation instance into three components: instruction, content, and response. The instruction specifies the judging context, the content consists of the long-form candidate outputs to be evaluated, and the response contains the judge's judgment. The following sections present the task formalization, describe task preparation, and summarize the resulting task statistics.

\subsection{Task Formalization}
\label{task-formalization}
We formalize LLM-as-a-judge as a function that takes an evaluation instruction and one or more candidate outputs as input, and produces a judgment as output. Each instance is represented as
\begin{equation}
    x_i = (\mathcal{I}_i, \mathcal{C}_i, z_i^{*}),
\end{equation}
where $\mathcal{I}_i$ denotes the instruction, $\mathcal{C}_i$ denotes the content to be evaluated, and $z_i^{*}$ denotes the reference judgment.

\textbf{Instruction.}
The instruction $\mathcal{I}$ denotes the judge-side context that guides the evaluation. We treat $\mathcal{I}$ as an extensible structured object instead of a fixed tuple, since different tasks may provide different forms of guidance. In \textit{LongJudgeBench}, an instruction typically contains the source query $q$, a judging protocol $m$, and a set of task-specific auxiliary fields $\mathcal{A}$:
\begin{equation}
    \mathcal{I} \doteq \{q, m\} \cup \mathcal{A}.
\end{equation}
Here, $q$ specifies the original generation task, and $m$ defines the expected form of judgment, such as pointwise scoring, pairwise comparison, or listwise ranking. The auxiliary fields $\mathcal{A}$ are optional and may include evaluation rubrics, criteria weights, reference materials, domain constraints, formatting requirements, or other task-specific metadata. 

\textbf{Content.}
The content $\mathcal{C}$ is the candidate output or set of candidate outputs to be evaluated. Depending on the judging protocol, $\mathcal{C}$ can take different forms:
\begin{equation}
    \mathcal{C} =
    \begin{cases}
        \{y_1\}, & \text{pointwise scoring}, \\
        \{y_1, y_2\}, & \text{pairwise comparison}, \\
        \{y_1, \ldots, y_K\}, & \text{listwise ranking}.
    \end{cases}
\end{equation}
Here, each $y_k$ is a long-form candidate output.

\textbf{Response.}
Given an instruction and content, an LLM judge $J_{\theta}$ produces a response:
\begin{equation}
    \hat{R}_i = J_{\theta}(\mathcal{I}_i, \mathcal{C}_i).
\end{equation}
The response may include both a natural-language rationale and a structured final judgment:
\begin{equation}
    \hat{R}_i = (\hat{e}_i, \hat{z}_i),
\end{equation}
where $\hat{e}_i$ is an optional explanation and $\hat{z}_i$ is the extracted judgment. The judgment format depends on the protocol. In pointwise evaluation, $\hat{z}_i$ is usually a scalar score or an ordinal label; in pairwise evaluation, $\hat{z}_i$ is a preference over two candidate outputs; and in listwise evaluation, $\hat{z}_i$ is a ranking or a set of scores over multiple outputs.

\begin{table*}[t]
\centering
\resizebox{\textwidth}{!}{%
\begin{tabular}{llcrrlrr}
\hline
Scenario                              & Data Source & Lang. & \# Instance & \# Output & Protocol  & \begin{tabular}[c]{@{}c@{}}Mean\\ Output Len.\end{tabular} & \begin{tabular}[c]{@{}c@{}}Median\\ Output Len.\end{tabular} \\ \hline
\multirow{2}{*}{Deep research}        & DR-Bench    & Zh    & 200       & 200       & Pointwise & 18,012.9 & 9,604.5  \\
                                      & RealDR      & Zh/En & 640       & 640       & Pointwise & 10,131.4 & 9,610.0  \\ \hline
Scientific survey                     & SurGE       & En    & 82        & 164       & Listwise  & 28,758.0 & 35,948.5 \\ \hline
Creative writing                      & WP-Bench    & Zh/En & 526       & 526       & Pairwise  & 3,508.9  & 2,178.5  \\ \hline
Long-chain analysis                  & Verify      & En    & 316       & 316       & Pointwise & 3,052.6  & 1,692.5  \\ \hline
Systematic review                     & MA         & En    & 180      & 120       & Pairwise  & 4,764.1  & 4,515.5  \\ \hline
Total                                 & --          & Zh/En & 1,944     & 1,966     & --        & 9,249.7  & --       \\ \hline
\end{tabular}%
}
\caption{Statistics of our \textit{LongJudgeBench}. In the following sections, we use DR-Bench, RealDR, WP-Bench, and Verify to denote DeepResearch Bench, RealDeepResearch, WritingPreferenceBench, and VerifyBench, respectively. Output length is measured in tokens using the \texttt{cl100k\_base} tokenizer from \texttt{tiktoken}.}

\label{tab:longjudgebench_stats}
\end{table*}

\subsection{Task Preparation}

\textit{LongJudgeBench} is designed to assess LLM judges across representative long-form output evaluation tasks. While LLM-as-a-judge has been adopted in many real-world long-form tasks, existing studies remain fragmented, often limited in scale and heterogeneous in data formats, judgment protocols, and evaluation targets. \textit{LongJudgeBench} consolidates these scattered settings into a unified benchmark, enabling systematic evaluation of judge reliability across diverse long-form scenarios. Specifically, we construct or collect six evaluation datasets from five real-world scenarios: (i) deep research, (ii) scientific survey, (iii) creative writing, (iv) long-chain analysis, and (v) systematic review. These datasets span multiple domains, two languages (Chinese and English), and three judgment protocols: pointwise, pairwise, and listwise evaluation. 


\textbf{Deep research.}
We first incorporate DeepResearch Bench~\cite{du2026deepresearch}, which provides expert-designed research questions and long research reports. However, its publicly released human-annotated subset is limited to 200 Chinese reports stored as plain strings, lacking multilingual coverage and omitting presentation-level information such as tables, figures, structured layouts, and other document elements commonly found in real reports. To address these limitations, we further construct and annotate RealDeepResearch, which contains 40 research tasks, 4 prompt styles, and 4 generation models, resulting in 640 long documents. RealDeepResearch covers both Chinese and English documents and preserves the original DOCX format, making it closer to real-world deep research evaluation. The annotations were carried out by expert annotators, who were compensated at a fair rate. The total cost was approximately USD 2,260. Further details are provided in Appendix~\ref{task_preparation_details}.

\textbf{Scientific survey.}
We include SurGE~\cite{su2026surge}, which provides 41 topics and 164 survey documents with human rankings over structural and content quality.

\textbf{Creative writing.}
We use a uniformly sampled subset of WritingPreferenceBench~\cite{ying2025beyond}, which focuses on pairwise preferences between long-form creative responses driven by subjective qualities such as style, creativity, and emotional resonance.

\textbf{Long-chain analysis.}
We include VerifyBench~\cite{yan2026verifybench}, where judges are given long-form model reasoning analyses and are asked to determine whether the final answer is correct.

\textbf{Systematic review.}
We include MA~\cite{xie2026benchmarking} for evaluating the insight dimension in clinical systematic reviews and meta-analyses.

Following the formalization in Section~\ref{task-formalization}, we normalize all datasets into a unified instance format while preserving their original evaluation protocols whenever possible. More details about task preparation can be found in Appendix~\ref{task_preparation_details}.

\subsection{Task Statistics}
After task preparation, \textit{LongJudgeBench} contains a diverse collection of long-form output evaluation instances across multiple scenarios and judging protocols. 
We report the statistics of the benchmark in Table~\ref{tab:longjudgebench_stats}. 
Compared with existing LLM-as-a-judge benchmarks, where candidate outputs are typically shorter than 500 tokens on average, \textit{LongJudgeBench} targets substantially longer outputs, with an average length exceeding 9,000 tokens. 
Beyond increasing output length, these instances expose LLM judges to real-world document-level evaluation demands, such as assessing overall organization, cross-section consistency, and coverage and depth across multiple subtopics.

\begin{table*}[t]
\centering
\resizebox{\textwidth}{!}{%
\begin{tabular}{llccccccccc}
\hline
\multirow{2}{*}{Model} 
& \multirow{2}{*}{Setting} 
& \multirow{2}{*}{DR-Bench} 
& \multirow{2}{*}{RealDR} 
& \multicolumn{2}{c}{SurGE} 
& \multirow{2}{*}{WP-Bench} 
& \multirow{2}{*}{Verify} 
& \multirow{2}{*}{MA} 
& \multirow{2}{*}{\begin{tabular}[c]{@{}c@{}}Avg.\\ w/o WP\end{tabular}} 
& \multirow{2}{*}{\begin{tabular}[c]{@{}c@{}}Avg.\\ w/ WP\end{tabular}} \\ 
\cline{5-6}
& & & & Structure & Content & & & & & \\ 
\hline

\multirow{4}{*}{\begin{tabular}[c]{@{}l@{}}Qwen3-32B\\(w/o thinking)\end{tabular}} 
& Vanilla     
& 0.5600 & 0.3929 & 0.5732 & 0.6179 & 0.6160 & 0.4968 & 0.5667 & 0.5346 & 0.5462 \\
& Rubric      
& 0.7033 & 0.3881 & 0.6016 & 0.6138 & 0.6065 & 0.4747 & 0.5278 & 0.5516 & 0.5594 \\
& Reference   
& 0.5867 & 0.3271 & 0.6260 & 0.7073 & -- & 0.7880 & 0.5111 & 0.5910 & 0.5910 \\
& Ref.+Rubric 
& 0.6833 & 0.4121 & 0.5935 & 0.6667 & -- & 0.7816 & 0.5111 & 0.6081 & 0.6081 \\
\hline

\multirow{4}{*}{\begin{tabular}[c]{@{}l@{}}Qwen3-32B\\(w/ thinking)\end{tabular}} 
& Vanilla     
& 0.5833 & 0.4144 & 0.3902 & 0.4553 & \textbf{0.6388} & 0.4589 & 0.5444 & 0.4744 & 0.4979 \\
& Rubric      
& 0.6967 & 0.5017 & 0.4472 & 0.3862 & 0.6369 & 0.4842 & 0.4222 & 0.4897 & 0.5107 \\
& Reference   
& 0.5367 & 0.4373 & 0.5203 & 0.5854 & -- & 0.7342 & 0.4722 & 0.5477 & 0.5477 \\
& Ref.+Rubric 
& 0.6800 & 0.5031 & 0.4309 & 0.4797 & -- & 0.6804 & 0.4944 & 0.5448 & 0.5448 \\
\hline

\multirow{4}{*}{Qwen3-Max} 
& Vanilla     
& 0.5500 & 0.4275 & 0.7073 & 0.7236 & 0.5875 & 0.5316 & 0.5389 & 0.5798 & 0.5809 \\
& Rubric      
& 0.7133 & 0.5019 & 0.8008 & 0.6382 & 0.5856 & 0.5443 & 0.5278 & 0.6211 & 0.6160 \\
& Reference   
& 0.6667 & 0.4808 & \textbf{0.8211} & 0.7317 & -- & 0.8291 & 0.5167 & \textbf{0.6744} & \textbf{0.6744} \\
& Ref.+Rubric 
& 0.7133 & 0.4683 & 0.4959 & 0.7561 & -- & 0.8133 & 0.5056 & 0.6254 & 0.6254 \\
\hline

\multirow{4}{*}{GPT-4o-mini} 
& Vanilla     
& 0.3833 & 0.3544 & 0.2276 & 0.2805 & 0.5114 & 0.4019 & 0.5611 & 0.3681 & 0.3886 \\
& Rubric      
& 0.6900 & 0.3458 & 0.2764 & 0.2317 & 0.5076 & 0.4051 & 0.5556 & 0.4174 & 0.4303 \\
& Reference   
& 0.6233 & 0.2435 & 0.4228 & 0.1707 & -- & 0.7975 & 0.5833 & 0.4735 & 0.4735 \\
& Ref.+Rubric 
& \textbf{0.7467} & 0.3208 & 0.5122 & 0.3577 & -- & 0.7943 & \textbf{0.6000} & 0.5553 & 0.5553 \\
\hline

\multirow{4}{*}{GPT-5.2}
& Vanilla     
& 0.4733 & 0.3554 & 0.4187 & 0.3293 & 0.5266 & 0.7089 & 0.4111 & 0.4495 & 0.4605 \\
& Rubric      
& 0.7100 & 0.4429 & 0.7154 & 0.2033 & 0.5722 & 0.6899 & 0.3889 & 0.5251 & 0.5318 \\
& Reference   
& 0.5633 & 0.5085 & 0.3089 & 0.2276 & -- & 0.8228 & 0.4444 & 0.4793 & 0.4793 \\
& Ref.+Rubric 
& 0.7033 & 0.4319 & 0.1545 & 0.2033 & -- & 0.8133 & 0.4222 & 0.4548 & 0.4548 \\
\hline

\multirow{4}{*}{DeepSeek-V4-Flash}
& Vanilla     
& 0.5933 & 0.4427 & 0.7073 & \textbf{0.7846} & 0.5837 & 0.5981 & 0.4611 & 0.5979 & 0.5958 \\
& Rubric      
& 0.7200 & \textbf{0.5710} & 0.6341 & 0.6057 & 0.5837 & 0.5570 & 0.3944 & 0.5804 & 0.5808 \\
& Reference   
& 0.6767 & 0.4067 & 0.7967 & 0.7398 & -- & \textbf{0.8924} & 0.4778 & 0.6650 & 0.6650 \\
& Ref.+Rubric 
& 0.7100 & 0.5552 & 0.7073 & 0.6341 & -- & 0.8133 & 0.4389 & 0.6431 & 0.6431 \\
\hline

\multirow{4}{*}{GLM-5.1}
& Vanilla     
& 0.5967 & 0.5067 & 0.6911 & 0.7520 & 0.5913 & 0.6171 & 0.5056 & 0.6115 & 0.6086 \\
& Rubric      
& 0.7067 & 0.5606 & 0.7276 & 0.6545 & 0.5779 & 0.5981 & 0.4778 & 0.6209 & 0.6147 \\
& Reference   
& 0.6900 & 0.5092 & 0.8130 & 0.6179 & -- & 0.8228 & 0.4444 & 0.6496 & 0.6496 \\
& Ref.+Rubric 
& 0.6967 & 0.5315 & 0.5610 & 0.6016 & -- & 0.7563 & 0.4722 & 0.6032 & 0.6032 \\
\hline

\multirow{4}{*}{Kimi-K2.6}
& Vanilla     
& 0.6433 & 0.4492 & 0.6463 & 0.7073 & 0.5589 & 0.5791 & 0.5389 & 0.5940 & 0.5890 \\
& Rubric      
& 0.7200 & 0.5175 & 0.7439 & 0.6179 & 0.5570 & 0.6203 & 0.5167 & 0.6227 & 0.6133 \\
& Reference   
& 0.6733 & 0.4577 & 0.5610 & 0.5610 & -- & 0.8481 & 0.5722 & 0.6122 & 0.6122 \\
& Ref.+Rubric 
& 0.6900 & 0.5056 & 0.4390 & 0.5691 & -- & 0.8196 & 0.5389 & 0.5937 & 0.5937 \\
\hline

\end{tabular}%
}
\caption{Accuracy of different LLM-as-a-judge baselines. SurGE reports structure and content quality. Ref.+Rubric denotes reference-based judging with rubric guidance. WP-Bench is unavailable for reference-based settings due to the absence of appropriate reference answers. Avg. w/o WP averages DR-Bench, RealDR, SurGE (Structure and Content), Verify, and MA. Avg. w/ WP further includes WP-Bench when available and ignores missing values.}
\label{tab:main-result}
\end{table*}

\section{Experimental Setup}
\subsection{Baselines}

We evaluate a broad set of LLM-as-a-judge baselines along two axes: the base model and the judging setting. This design allows us to analyze how judge reliability varies with model capability, the available evaluation information, and their interaction under long-form evaluation demands.

\textbf{Base models.}
We instantiate LLM judges using models from representative model families,
including Qwen3-32B (w/o thinking), Qwen3-32B
(w/ thinking), Qwen3-Max~\cite{yang2025qwen3}, GPT-4o-mini, GPT-5.2~\cite{singh2025openai}, DeepSeek-V4-Flash~\cite{xu2026deepseek}, GLM-5.1~\cite{zeng2026glm}, and Kimi-K2.6~\cite{team2025kimi}. These models cover both
proprietary and open-weight systems, and represent different strengths in
instruction following, long-context understanding, and reasoning. All judges are evaluated using the same task inputs and output-extraction procedure. We set the temperature to 0 for all runs to ensure a fair comparison across base models.

\textbf{Judging settings.}
For each base model, we evaluate several judging settings that differ in whether they provide external evaluation anchors, such as rubrics or references.

\textbf{Vanilla prompting} asks the model to directly evaluate the candidate output and produce a final judgment. This represents the simplest and most commonly used LLM-as-a-judge setting.

\textbf{Rubric-based judging} provides the judge with task-specific rubrics or criteria. The judge is asked to assess the candidate output according to these criteria before producing the final judgment.

\textbf{Reference-based judging} provides the judge with additional reference information, such as reference answers, expert-written reports, or task-specific evidence. The judge is expected to use this information as an evaluation anchor when assessing the candidate output.

\textbf{Reference+Rubric judging} combines both types of auxiliary information. 
The judge is given reference information as comparison anchors and rubrics as evaluation criteria. 
This setting examines whether explicit references and structured criteria can jointly improve judgment reliability.


\subsection{Metrics}

We evaluate judge reliability by measuring the agreement between model-produced judgments and reference judgments using accuracy, Spearman's rank correlation, and Kendall's $\tau_b$. Unless otherwise specified, each metric is computed within a query group and then averaged across groups. Accuracy is used as the main metric in our experiments.

\textbf{Accuracy.}
We compute accuracy as the agreement between predicted judgments and reference judgments. Depending on the dataset, the judgment unit is either a preference decision or a correctness decision. For preference-based datasets, we convert both model judgments and reference judgments into pairwise preferences. For a query $q$ with candidate outputs $\{y_{q,1}, \ldots, y_{q,K}\}$, let $\mathcal{P}_q$ denote the set of all valid candidate pairs. For each pair $(a,b) \in \mathcal{P}_q$, let $z^{*}_{q,a,b}$ and $\hat{z}_{q,a,b}$ denote the reference and predicted preferences, respectively. The predicted preference is obtained directly in pairwise evaluation, by comparing candidate scores in pointwise evaluation, and by comparing candidate ranks in listwise evaluation. We compute preference accuracy as
\begin{equation}
    \mathrm{Acc}_q
    =
    \frac{1}{|\mathcal{P}_q|}
    \sum_{(a,b) \in \mathcal{P}_q}
    \mathbb{I}[\hat{z}_{q,a,b} = z^{*}_{q,a,b}],
\end{equation}
where a prediction is considered correct only when it exactly matches the reference preference, including cases where both the reference and prediction indicate a tie. To reduce position bias in pairwise evaluation, we evaluate each pair in both answer orders and average the two results.

For correctness-based datasets such as VerifyBench, the reference label indicates whether the final answer is correct. We therefore compute accuracy directly as label agreement over all instances.

\textbf{Spearman's rank correlation.}
Given reference scores $\mathbf{s}^{*}_q = (s^{*}_{q,1}, \ldots, s^{*}_{q,K})$
and judge scores $\hat{\mathbf{s}}_q = (\hat{s}_{q,1}, \ldots, \hat{s}_{q,K})$
for query $q$, we first convert them into ranks:
\begin{equation}
    r^{*}_{q,k} = \operatorname{rank}(s^{*}_{q,k}),
    \qquad
    \hat{r}_{q,k} = \operatorname{rank}(\hat{s}_{q,k}).
\end{equation}
Spearman's rank correlation is computed as
\begin{equation}
    \rho_q =
    \frac{
    \sum_{k=1}^{K}
    (\hat{r}_{q,k} - \bar{\hat{r}}_q)
    (r^{*}_{q,k} - \bar{r}^{*}_q)
    }
    {
    \sqrt{
    \sum_{k=1}^{K}
    (\hat{r}_{q,k} - \bar{\hat{r}}_q)^2
    }
    \sqrt{
    \sum_{k=1}^{K}
    (r^{*}_{q,k} - \bar{r}^{*}_q)^2
    }
    },
\end{equation}
where $\bar{\hat{r}}_q$ and $\bar{r}^{*}_q$ denote the mean ranks of the judge
scores and reference scores, respectively.

\textbf{Kendall's rank correlation.}
We report Kendall's $\tau_b$ to account for tied rankings. For each query $q$, we compare the pairwise ordering induced by the judge scores $\hat{s}$ with that induced by the reference scores $s^*$. Let $C_q$ and $D_q$ be the numbers of concordant and discordant candidate pairs, respectively, and let $T_{\hat{s},q}$ and $T_{s^*,q}$ be the numbers of tied pairs that occur only in the judge scores and only in the reference scores. We compute
\begin{equation}
    \tau_{b,q} =
    \frac{C_q - D_q}
    {\sqrt{(C_q + D_q + T_{\hat{s},q})(C_q + D_q + T_{s^*,q})}} .
\end{equation}

\section{Experimental Results}

\subsection{Main Results}

Table~\ref{tab:main-result} reports the accuracy of different judging settings. Due to space limitations, we focus on accuracy in the main results and provide additional metrics in Appendix~\ref{more-results}. We use Avg. w/ WP as the default aggregate score unless otherwise specified. The main observations are summarized below.

\textbf{Overall, \textit{LongJudgeBench} remains highly challenging.}
Existing LLM-as-a-judge methods achieve only moderate performance on \textit{LongJudgeBench}. The best configuration, Qwen3-Max with reference, reaches an average accuracy of 0.6744, followed by DeepSeek-V4-Flash + Reference at 0.6650 and GLM-5.1 + Reference at 0.6496. Across all 32 model--setting combinations, only 12 exceed 0.60, and the overall mean is 0.5639, only slightly above the random pairwise baseline of 0.5. These results indicate that reliable long-form judging remains far from solved.

\textbf{Judge performance varies substantially across models and scenarios.}
Qwen3-Max, DeepSeek-V4-Flash, and GLM-5.1 achieve the highest aggregate scores among the evaluated judges, while GPT-4o-mini and GPT-5.2 obtain lower scores. This suggests that stronger general-purpose capabilities may not always translate into stronger long-form judging performance. Scenario-level results also vary widely across tasks. Judges tend to perform better on tasks with clearer evaluation anchors: Verify achieves the highest average score of 0.6742, likely because it mainly requires checking whether the final answer matches the reference; DR-Bench also obtains a relatively high score of 0.6464, where judges evaluate plain-text research reports with more explicit quality criteria. In contrast, performance is lower on scenarios that may require broader document understanding or more subjective judgment: RealDR reaches 0.4460, possibly reflecting the added difficulty of realistic document structure and multilingual content; MA obtains 0.4983, as medical insight evaluation requires domain-specific evidence understanding; and WP-Bench remains challenging at 0.5776, as creative-writing evaluation often depends on subjective preferences. This contrast suggests that the difficulty of long-form judging depends not only on output length, but also on how explicit, structured, and objective the evaluation target is.

\textbf{References and rubrics help, but their benefits are not universal.}
Among the four judging settings, Reference achieves the best average performance, improving the overall score from 0.5334 under Vanilla to 0.5866. The gain is particularly clear on tasks with explicit evaluation anchors, such as Verify, where the average score increases from 0.5491 to 0.8169. Rubric also improves over Vanilla, especially for multidimensional research-report evaluation: DR-Bench increases from 0.5479 to 0.7075, and RealDR increases from 0.4179 to 0.4787. However, these improvements are not uniform across scenarios. Rubric underperforms Vanilla on SurGE-Content and MA, while Reference brings only limited gains on RealDR. Moreover, the benefit of combining reference and rubric guidance is not simply additive: Ref.+Rubric reaches 0.5786, slightly below Reference alone at 0.5866. These results suggest that the Reference and Rubric may offer useful evaluation signals, but effectively integrating them remains an important challenge for reliable long-form judging.

\textbf{Thinking mode shows mixed effects on Qwen3-32B as a judge.}
For Qwen3-32B, enabling thinking mode does not lead to consistent gains across scenarios. It improves RealDR from 0.3801 to 0.4641 and brings a modest gain on WP-Bench from 0.6113 to 0.6379. However, these improvements are not reflected in the overall average, which decreases from 0.5762 without thinking to 0.5253 with thinking. Overall, thinking mode may be helpful in certain scenarios, but its effect is highly task-dependent rather than always beneficial.

\subsection{Output Length Sensitivity Analysis}
In this section, we analyze how output length affects the LLM judges' reliability. To reduce confounding from additional judging inputs, we conduct this analysis under the Vanilla setting, without references or rubrics. The analysis covers four strong judge models, Qwen3-Max, DeepSeek-V4-Flash, GLM-5.1, and Kimi-K2.6, on all six datasets. For preference-based datasets, we measure output length by the average token length of the two candidate outputs; for correctness-based datasets such as Verify, we use the token length of the evaluated output. For each dataset, we sort instances by output length and split them into four equal-sized quantile bins, Q1--Q4, from shortest to longest. We then compute accuracy within each bin. We use adaptive bins because length distributions differ substantially across datasets, making fixed intervals prone to empty or highly imbalanced bins.

\begin{figure}[t]
    \centering
    \includegraphics[width=\columnwidth]{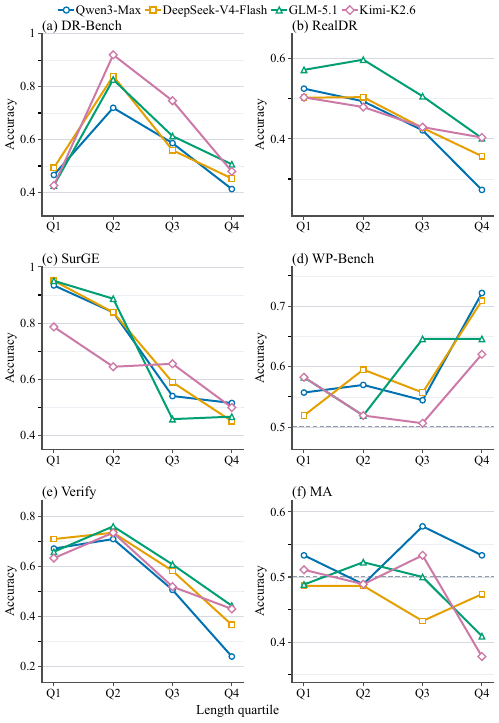}
    \caption{Output length sensitivity of LLM judges across six datasets. For SurGE, we plot only Structure since Content shows a similar trend.}
    \label{fig:2}
\end{figure}

As shown in Figure~\ref{fig:2}, we analyze the effect of output length on judge performance from two perspectives: the direction of performance changes and their consistency across models. Overall, the relationship between output length and judge accuracy is dataset-specific and cannot be simply summarized as \textit{longer is worse}. RealDR, SurGE, and Verify generally show decreasing trends from Q1 to Q4, and these trends are relatively consistent across judge models. In contrast, DR-Bench exhibits a non-monotonic pattern, with all four models achieving higher accuracy around Q2 than on Q1 and Q4. WP-Bench does not show a decreasing trend with longer outputs; some models instead achieve higher accuracy on longer bins. MA also lacks a clear monotonic pattern, with inconsistent directions across models. These results support our observation that long-form evaluation is not merely a matter of output length, but also involves task-specific document-level demands, such as subjective preference assessment in WP-Bench and domain-specific medical insight evaluation in MA.

\subsection{Case Study and Failure Analysis}
In this section, we further summarize the challenges of long-form evaluation revealed by \textit{LongJudgeBench} and discuss potential improvements.

\textbf{Misled by superficial coverage.} Long-form outputs are often lengthy, well-structured, and rich in background knowledge, which can create an impression of sufficient coverage without truly addressing the user’s core question. In a \textit{computational chemistry} case, the user asks how to simulate an \textit{external electric field} in \textit{Gaussian} for a \textit{single-atom catalyst} system with uncertain \textit{molecular orientation}. The key issue is not general \textit{external-field simulation}, but how to handle the \textit{field direction} under \textit{orientation uncertainty}. Although the candidate response contains about 12K tokens and extensively discusses \textit{Gaussian Field syntax}, \textit{VASP}, and other general topics, it devotes only a few sentences to the core issue. Still, Qwen3-Max assigns an overall score of 9.26, while the human score is only 5.87. This case shows that LLM judges can mistake surface richness for substantive answers. A more reliable long-form judge should look beyond coverage and structure to assess whether the output truly addresses the user’s central need.

\textbf{Scenario-specific concept misgrounding.} Long-form generation tasks often involve complex \textit{domain-specific concepts}, but the LLM judge may lack the \textit{conceptual discrimination} needed for reliable evaluation. In a \textit{technical survey} case, the user asks about the concrete implementation of Anthropic’s recently released \textit{Streamable HTTP}. Although the candidate response is about 14K tokens long, it mainly discusses general \textit{SSE-based streaming} in the \textit{Anthropic Messages API}. The Qwen3-Max judge incorrectly treats \textit{Messages API streaming} and \textit{Streamable HTTP} as the same concept, assigning an overall score of 8.85, while the human score is only 5.37. Similarly, Table~\ref{tab:main-result} further shows that LLM judges perform poorly on the MA medical-domain dataset, suggesting that it may struggle to apply domain-specific evaluation criteria in specialized scenarios. Future long-form judges need stronger domain concept disambiguation and better adaptation to different evaluation demands.

\textbf{Other failures.} LLM judges also exhibit some systematic biases and practical failures. As detailed in Appendix~\ref{Analysis-Details}, their decisions can be affected by irrelevant factors such as candidate output order, indicating \textit{position bias}. In addition, some judgments fail because of \textit{context-window overflow} or \textit{safety-policy rejection}, especially when long outputs are combined with references and rubrics or when inputs contain sensitive legal-domain content. These findings show that long-form evaluation is constrained not only by judgment accuracy, but also by robustness to bias, prompt complexity, and model safety behavior.


\section{Conclusion}
In this paper, we introduce \textit{LongJudgeBench}, a comprehensive benchmark for evaluating LLM-as-a-judge reliability in long-form output evaluation. Covering diverse real-world scenarios and evaluation protocols, \textit{LongJudgeBench} enables systematic analysis of LLM judges across different settings. Our experiments show that long-form judging remains highly challenging, and we hope this work will help pave the way for future research on reliable LLM-as-a-judge methods.

\section{Limitations}
We acknowledge several limitations in this study and aim to address them in future work. First, although \textit{LongJudgeBench} covers multiple representative long-form evaluation scenarios, it still cannot fully capture the diversity of real-world long-form generation tasks. We plan to further expand the benchmark with more domains, task types, and application settings. Second, our experiments consider several representative LLM judges and judging settings, but more advanced judge designs, such as retrieval-augmented judging, multi-agent judging, and calibration-based methods, remain to be explored. We hope future work can build upon \textit{LongJudgeBench} to develop more reliable and robust long-form output evaluation methods.

\section{Ethical Considerations}
Throughout this research, ethical considerations have been integral to ensuring the responsible development and application of AI technologies. We are committed to the principles of open research and the scientific value of reproducibility. Therefore, we have made all the code from our study publicly accessible on GitHub. This transparency allows the community to validate our results and promotes the use of our methods in various settings. Moreover, we used AI assistants only for language polishing and editing to improve clarity, grammar, and readability.

\section*{Acknowledgments}
This work is supported by the Research Project of Quancheng Laboratory, China (Grant No. QCL20250105).

\bibliography{main}

\appendix
\section{License}

In this section, we clarify the copyright and licensing status of
\textit{LongJudgeBench} to ensure that users can utilize the benchmark in a legal and compliant manner.

\textit{LongJudgeBench} is constructed from open-source datasets and data collected
and annotated by us. All external datasets used in \textit{LongJudgeBench}
are publicly available, and we have carefully reviewed their licenses and
usage terms to ensure that their inclusion in the benchmark is permitted.
For our self-collected data, annotations, metadata, model outputs, and
evaluation protocols, we hold the rights to release them as part of the
benchmark. While the processed benchmark is released for public use, the copyright of
the original third-party resources remains with their respective authors or
organizations. Users are expected to respect the original licenses,
citation requirements, and terms of use associated with these resources
when using \textit{LongJudgeBench}.

We release our code, self-collected data, annotations, metadata, model outputs, and evaluation protocols under the MIT License where permitted. 
Third-party datasets included in \textit{LongJudgeBench} remain subject to their original licenses and terms of use. 
Users should follow the corresponding licenses, citation requirements, and usage restrictions of the original resources. If you believe that
\textit{LongJudgeBench} contains any content that infringes upon your rights or
violates the license terms of an original resource, please contact us at
any time, and we will promptly review and address the issue.

\section{Analysis Details}
\label{Analysis-Details}
In this section, we further analyze some systematic biases and practical failures in \textit{LongJudgeBench}.

\textbf{Position bias.}
For pairwise preference-based datasets, a reliable judge should give consistent decisions regardless of the order in which candidate outputs are presented. To measure position bias, we swap the order of the two candidate outputs and compare the resulting preference with the original judgment. We define an inconsistent pair as a pair where the judge gives contradictory content-level preferences after the order swap, which corresponds to selecting the same displayed position in both orders. A higher inconsistency rate therefore indicates stronger sensitivity to candidate order.

Table~\ref{tab:position-bias} reports the inconsistency rates under the Vanilla setting on WP-Bench and MA. The results show that position bias varies substantially across models and datasets. GPT-4o-mini shows an especially high inconsistency rate of 78.7\% on WP-Bench, suggesting strong dependence on candidate order. GPT-5.2 also shows high inconsistency on MA, reaching 55.6\%. In contrast, Kimi-K2.6 and Qwen3-32B without thinking exhibit lower inconsistency rates on both datasets. These results suggest that some LLM judges may rely on positional cues rather than stable quality criteria, especially when comparing long-form candidate outputs.

\begin{figure}[t]
    \centering
    \includegraphics[width=\columnwidth]{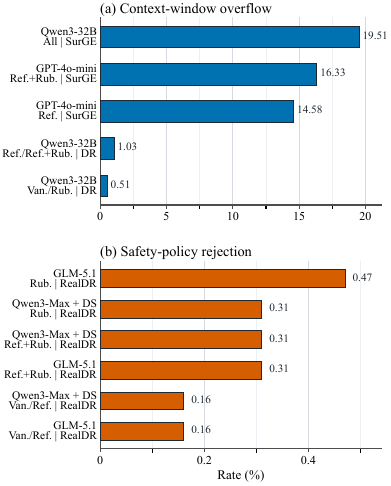}
    \caption{
    Special failure rates of LLM judges in long-form evaluation. The rate denotes the percentage of evaluation instances where the judge fails to produce a valid judgment due to context-window overflow or safety-policy rejection. DS denotes DeepSeek-V4-Flash. When multiple model names are joined by “+”, they share the same failure rate under the corresponding setting and dataset.
    }
    \label{fig:failure-rates}
\end{figure}

\textbf{Practical failures.}
LLM judges also encounter practical failures that prevent valid scoring. As shown in Figure~\ref{fig:failure-rates}, these failures are concentrated in a few model--setting--dataset groups. \textit{Context-window overflow} occurs when long prompts exceed the model's input limit, especially when outputs, references, and rubrics are combined. \textit{Safety-policy rejection} occurs when judges refuse to score certain inputs, mainly in RealDR queries involving criminal-law-related content. Since these failures prevent valid judgments, we treat them as incorrect when computing accuracy. These results show that long-form evaluation is also constrained by context length, prompt complexity, and model safety behavior.

\begin{table}[t]
\centering
\begin{tabular}{lcc}
\hline
Model & WP-Bench & MA \\
\hline
GLM-5.1 & 17.9\% & 33.3\% \\
GPT-4o-mini & \textbf{78.7\%} & 48.9\% \\
GPT-5.2 & 46.0\% & \textbf{55.6\%} \\
DeepSeek-V4-Flash & 62.0\% & 45.6\% \\
Qwen3-Max & 43.0\% & 38.9\% \\
Qwen3-32B & 21.3\% & 37.8\% \\
Qwen3-32B w/o thinking & 16.0\% & 21.1\% \\
Kimi-K2.6 & 10.6\% & 23.3\% \\
\hline
\end{tabular}
\caption{Position bias analysis under the Vanilla setting on WP-Bench and MA. We report the inconsistency rate after swapping the order of the two candidate outputs. Higher values indicate stronger position bias.}
\label{tab:position-bias}
\end{table}

\section{Task Preparation Details}
\label{task_preparation_details}

This section provides additional details about the datasets used in
\textit{LongJudgeBench}, including their sources, task formats, annotation protocols,
and adaptation procedures. Following Section~\ref{task-formalization}, we
normalize all datasets into a unified schema while preserving their original
evaluation protocols whenever possible.

\subsection{Deep Research}

\paragraph{DeepResearch Bench.}
DeepResearch Bench~\cite{du2026deepresearch} is a benchmark for evaluating
deep research agents (DRAs). \textit{LongJudgeBench} uses its open-source Chinese
subset and the corresponding human annotations to evaluate the reliability of
LLM-as-a-judge on long research reports. Each task contains a PhD-level research question designed by domain experts.
The full benchmark covers 22 domains, such as science and technology, finance
and business, software and Internet, art and design, history, transportation,
and travel. In our setting, we use 50 Chinese research tasks. Four deep
research agents, including openai-deepresearch, gemini-2.5-pro-deepresearch,
grok-deeper-search, and perplexity-Research, generate one report for each
task, resulting in 200 reports. The report lengths range from about 2,000 to
158,000 tokens, with a median length of about 9,605 tokens.

For reference judgments, three experts with relevant research
backgrounds independently score each report on a 0--100 scale along four
dimensions: comprehensiveness, insight, instruction following, and
readability. Comprehensiveness measures whether the report covers the key
aspects and information required by the task. Insight evaluates analytical
depth and the novelty of viewpoints. Instruction following measures how
precisely the report follows the explicit task requirements. Readability
evaluates fluency, clarity, and ease of understanding. The final human score is computed as the
weighted sum of the four dimension-level scores. The original paper reports an inter-annotator agreement of 68.44\%, and we use the average score across all human annotations as the reference judgment.

\paragraph{RealDeepResearch.}
RealDeepResearch is our self-constructed multilingual long-form report evaluation dataset. 
It is designed to complement DeepResearch Bench by adding multilingual coverage and preserving realistic document-level information, such as tables, figures, and layouts. 
Unlike datasets that evaluate only plain-text reports, RealDeepResearch focuses on realistic document-level qualities, including logical organization, presentation, and bias-related issues.

RealDeepResearch contains 40 tasks, including 29 Chinese and 11 English tasks. 
These tasks cover 12 disciplines, including education, arts, economics, computer science, medicine, sociology, electronic engineering, environmental engineering, law, history, philosophy, and mechanical engineering. 
Its key design is to combine multiple prompt styles with multiple generation models, thereby creating diverse long-document quality evaluation scenarios. For each task, we design four prompt styles: high-quality prompts with clear evaluation expectations, weakened prompts that partially weaken but do not remove key requirements, bias-inducing prompts that introduce superficial authority signals and hidden logical flaws, and underspecified prompts without clear scoring criteria. 
Each prompt is answered by four generation models, namely Gemini, Grok, Mita, and Perplexity. 
This produces 16 documents per task and 640 documents in total. 
The median document length is about 9,610 tokens.

For reference judgments, we recruit two annotators with relevant backgrounds for each task--document pair.
They independently score each document on a 0--10 scale along three dimensions: logical structure, presentation form, and bias checking. 
These annotations focus on document-level quality rather than exhaustive domain-specific factual verification. 
Logical structure evaluates the overall organization, section division, heading-content alignment, and reasoning coherence. 
Presentation form evaluates language fluency, conciseness, readability, and coordination between text and visual elements. 
Bias checking evaluates fairness, logical fallacies, one-sidedness, over-reliance on authority, and misleading statements. 
The three dimensions have document-specific weights specified in the task annotation guideline, and the final score is computed as their weighted sum. 
If the two annotators have a large disagreement according to the predefined adjudication rule, a third annotator is introduced for adjudication. 
The annotation agreement is measured by the average-measure ICC, with $ICC_{avg} = 0.76$, indicating high agreement. We use the average score across all human
annotations as the reference judgment.

\subsection{Scientific Survey}

\paragraph{SurGE.}
SurGE~\cite{su2026surge} is a benchmark and evaluation framework for
automatic scientific survey generation. \textit{LongJudgeBench} uses 41 manually
annotated survey topics from SurGE to evaluate whether LLM judges can recover
human rankings over long survey documents.

Each topic corresponds to a computer science survey topic and contains four
survey documents: GT, a high-quality human-written survey with an average
length of 23,419 tokens; AutoSurvey, an automated survey generation pipeline
with an average length of 47,028 tokens; ID, an iterative refinement-based
survey generation system with an average length of 42,429 tokens; and Naive, a
simple RAG baseline with an average length of 2,156 tokens. The large length
gap across systems, especially between Naive and AutoSurvey, provides a useful
setting for examining whether judges are overly sensitive to output length.

For reference judgments, the annotation team consists of doctoral researchers in computer science with prior experience in academic peer review. For each test query, two experts independently rank the four anonymous surveys on a 1–4 scale, where 1 indicates the best, and 4 indicates the worst. The rankings are produced along two dimensions: structural quality and content quality. Structural quality evaluates the organization, hierarchy, and logical structure of the survey, while content quality assesses factual accuracy, coverage, and depth.

\subsection{Creative Writing}

\paragraph{WritingPreferenceBench.}
WritingPreferenceBench~\cite{ying2025beyond} is a benchmark for evaluating
cross-cultural subjective writing preferences. \textit{LongJudgeBench} uses a subset of
263 human-validated preference pairs, covering 8 creative writing domains:
fictional narrative, non-fiction writing, functional documents, promotional
communication, poetry, scriptwriting, role-playing, and humor. These pairs
span 51 fine-grained genres. The dataset controls for objective confounders such as grammatical errors, factual mistakes, and length differences, so that the evaluation focuses on subjective writing quality.

Each instance contains a writing instruction and two anonymized responses,
denoted as Response A and Response B. Both responses are verified to be
grammatically correct, factually accurate, and length-matched. The responses
are generated by more than 20 models, including Claude, GPT, Gemini, Doubao,
and DeepSeek, with five samples per query at temperature 0.8. The median
document length in our subset is about 2,179 tokens.

The reference preferences are constructed through a multi-stage human
annotation pipeline. First, responses with objective defects are automatically
filtered out, accounting for about 15\% of the data. Then, 11 professional
writing evaluators undergo an 8-hour rubric calibration process and score each
response on a 0--3 scale along three dimensions: creativity, style, and
emotional resonance. Creativity measures originality and imagination. Style
captures linguistic expression, narrative techniques, and rhetorical quality.
Emotional resonance measures whether the response can evoke convincing
emotional engagement. For quality control, each pair must have at least two
out of three annotators agreeing on the preference, and the minimum score gap
must satisfy $\Delta \geq 1$. The higher-scoring response is labeled as
chosen, and the lower-scoring response is labeled as rejected. Pairs with tied
ground-truth scores are excluded from evaluation.

\subsection{Long-Chain Analysis}

\paragraph{VerifyBench.}
VerifyBench~\cite{yan2026verifybench} is a benchmark for evaluating
reference-based reward systems. \textit{LongJudgeBench} samples 316 instances from its
VerifyBench-Hard subset, where judges need to determine whether a model output
is consistent with a reference answer.

VerifyBench collects diverse instruction-answer pairs from multiple open
datasets and covers a broad range of knowledge domains. It uses 22 open-source
and closed-source LLMs to generate responses. Each instance contains an
instruction, a model response, a reference answer, and metadata including
answer type, answer subtype, and source. These metadata support fine-grained breakdown analysis across different verification settings.

The reference judgment is a binary correctness label, denoted by
gold\_correct, indicating whether the model response matches the reference
answer. Each instance is independently annotated by at least two human
annotators. The annotations are checked through a strict quality-control
process. VerifyBench-Hard is further constructed from cases where strong
models show substantial disagreement, followed by comprehensive human
annotation to ensure label quality.

\subsection{Systematic Review}

\paragraph{MA.}
MA~\cite{xie2026benchmarking} is a benchmark for evaluating generated systematic review and
meta-analysis reports. It is based on 88 clinical systematic review and
meta-analysis papers, covering medical areas such as oncology, cardiovascular
disease, and endocrinology. Each task requires an LLM to generate a complete
systematic review report according to the clinical research question under the
PICO framework.

Each paper contains the title, PICO elements, inclusion and exclusion
criteria, effect direction, and a summary of key insights. The PICO elements
include topic, population, intervention, comparison condition, and outcome.
The effect direction is labeled as Positive, Negative, or Mixed.

Six generation systems produce candidate reports by combining different base
models and retrieval methods: gpt5\_bm25, gpt5\_dense, glm5\_bm25,
dsr1\_bm25, auto\_bm25, and auto\_dense. Specifically, gpt5\_bm25 uses
GPT-Researcher with GPT-5 and BM25; gpt5\_dense uses GPT-Researcher with
GPT-5 and Dense-FT; glm5\_bm25 uses GPT-Researcher with GLM-5 and BM25;
dsr1\_bm25 uses GPT-Researcher with DeepSeek-R1 and BM25; auto\_bm25 uses
AutoMeta with GPT-5 and BM25; and auto\_dense uses AutoMeta with GPT-5 and
Dense-FT.

For reference judgments, the authors selected 30 papers from the 88 papers for complete annotation. 
The annotations were completed by eight graduate students with strong English proficiency and backgrounds in medicine or biology. They were recruited through an open call, screened by the laboratory, and compensated at approximately RMB 100 per hour, with a total annotation cost of about RMB 58,000. 
Each comparison was independently evaluated by all eight annotators in a pairwise manner along the insight dimension, which evaluates analytical depth and key findings. Annotators use a five-level preference scale: A2, A1, tie, B1, and B2. A2 and B2 indicate strong preferences, A1 and B1 indicate mild preferences, and tie indicates no clear preference. The insight dimension covers three fixed system pairs: gpt5\_bm25 vs. auto\_bm25, gpt5\_bm25 vs. gpt5\_dense, and auto\_bm25 vs. auto\_dense, with 30 papers for each pair and 90 comparisons in total. Since we evaluate each pair in both answer orders to reduce position bias, MA contributes 180 pairwise evaluation instances to \textit{LongJudgeBench}. To aggregate the annotations, we map the five-level preference labels A2, A1, tie, B1, and B2 to ordinal scores 0, 1, 2, 3, and 4, respectively. For each pairwise instance, we average the scores from the eight annotators to obtain a reference preference score. The final reference judgment is assigned to A if the averaged score is below 2, to B if it is above 2, and to a tie otherwise.

\subsection{Normalization Across Datasets}

We normalize all datasets into the unified \textit{LongJudgeBench} format. Pointwise
datasets, including DeepResearch Bench, RealDeepResearch, and VerifyBench,
contain one candidate output with a human score or correctness label. Pairwise
datasets, including WritingPreferenceBench and MA, contain two candidate
outputs with a human preference label. The listwise dataset, SurGE, contains
multiple candidate outputs with human rankings.

For pairwise evaluation, we present each candidate pair in both answer orders to reduce position bias. Across all datasets,
we preserve the original instructions, outputs, references, rubrics, and
metadata whenever available. This normalization enables \textit{LongJudgeBench} to
evaluate LLM-as-a-judge across diverse long-form scenarios while
maintaining comparability across different judgment protocols.

\section{Prompt Details}
In this section, we present the key prompts used in
\textit{LongJudgeBench}. Table~\ref{prompt:deepresearch-vanilla} shows one representative vanilla pointwise prompt for DeepResearch Bench. 
Full prompts, output schemas, and extraction rules are provided in the released code.

\section{More Evaluation Results}
\label{more-results}
Tables~\ref{tab:spearman-result} and~\ref{tab:kendall-result} report the Spearman and Kendall rank correlations of different LLM judges. Compared with accuracy, these two metrics provide a complementary view by measuring whether model judgments preserve the relative ranking of candidate responses. We observe several differences from the accuracy-based results. First, the highest average rank correlations are achieved by Qwen3-32B (w/o thinking) with rubric guidance, reaching 0.5647 Spearman and 0.4667 Kendall, narrowly outperforming DeepSeek-V4-Flash under the vanilla setting (0.5570 and 0.4663, respectively). In contrast, the best accuracy is obtained by reference-based settings such as Qwen3-Max with Reference. This suggests that a judge can be good at preserving relative response rankings even when it is not the strongest in pairwise decisions. Second, the effects of additional evaluation information are highly model-dependent. Rubric guidance often improves rank correlation, whereas references provide less consistent gains, and combining references with rubrics frequently reduces correlation, particularly on SurGE. Third, the thinking variant of Qwen3-32B does not show clear advantages over the non-thinking variant, and often obtains lower rank correlations, suggesting that longer reasoning does not necessarily lead to more stable ranking behavior.

Overall, Spearman and Kendall reveal useful aspects of ranking consistency that are not fully captured by accuracy. However, since these rank-based metrics are not suitable for all evaluation paradigms in \textit{LongJudgeBench}, we recommend using accuracy as the primary metric, with Spearman and Kendall serving as auxiliary analyses.

\begin{table*}[t]
\centering
\resizebox{\textwidth}{!}{%
\begin{tabular}{llcccccccc}
\hline
\multirow{2}{*}{Model} 
& \multirow{2}{*}{Setting} 
& \multirow{2}{*}{DR-Bench} 
& \multirow{2}{*}{RealDR} 
& \multicolumn{2}{c}{SurGE} 
& \multirow{2}{*}{WP-Bench} 
& \multirow{2}{*}{Verify} 
& \multirow{2}{*}{MA} 
& \multirow{2}{*}{Avg.} \\ 
\cline{5-6}
& & & & Structure & Content & & & & \\ 
\hline

\multirow{4}{*}{\begin{tabular}[c]{@{}l@{}}Qwen3-32B\\(w/o thinking)\end{tabular}} 
& Vanilla     & 0.3278 & 0.3394 & 0.6582 & 0.7562 & -- & -- & -- & 0.5204 \\
& Rubric      & 0.4697 & 0.2376 & 0.7567 & 0.7946 & -- & -- & -- & \textbf{0.5647} \\
& Reference   & 0.4001 & 0.2582 & 0.6719 & 0.8179 & -- & -- & -- & 0.5370 \\
& Ref.+Rubric & 0.4385 & 0.2739 & 0.6004 & 0.7491 & -- & -- & -- & 0.5155 \\
\hline

\multirow{4}{*}{\begin{tabular}[c]{@{}l@{}}Qwen3-32B\\(w/ thinking)\end{tabular}} 
& Vanilla     & 0.4251 & 0.2345 & 0.3119 & 0.4878 & -- & -- & -- & 0.3648 \\
& Rubric      & 0.4973 & 0.2654 & 0.6505 & 0.3914 & -- & -- & -- & 0.4512 \\
& Reference   & 0.3780 & 0.2591 & 0.5890 & 0.7117 & -- & -- & -- & 0.4845 \\
& Ref.+Rubric & 0.4316 & 0.2331 & 0.4789 & 0.5811 & -- & -- & -- & 0.4312 \\
\hline

\multirow{4}{*}{Qwen3-Max} 
& Vanilla     & 0.4140 & 0.3497 & 0.5224 & 0.5939 & -- & -- & -- & 0.4700 \\
& Rubric      & \textbf{0.5468} & 0.3327 & 0.6691 & 0.4171 & -- & -- & -- & 0.4914 \\
& Reference   & 0.4756 & \textbf{0.3813} & 0.7073 & 0.6310 & -- & -- & -- & 0.5488 \\
& Ref.+Rubric & 0.4879 & 0.3270 & 0.1985 & 0.6315 & -- & -- & -- & 0.4112 \\
\hline

\multirow{4}{*}{GPT-4o-mini} 
& Vanilla     & 0.2785 & 0.3537 & -0.2032 & 0.0187 & -- & -- & -- & 0.1119 \\
& Rubric      & 0.4115 & 0.3780 & -0.0037 & -0.1183 & -- & -- & -- & 0.1669 \\
& Reference   & 0.4525 & 0.2394 & -0.0640 & -0.5657 & -- & -- & -- & 0.0156 \\
& Ref.+Rubric & 0.4942 & 0.3084 & 0.1444 & -0.0527 & -- & -- & -- & 0.2236 \\
\hline

\multirow{4}{*}{GPT-5.2}
& Vanilla     & 0.2565 & 0.1825 & 0.1355 & -0.0683 & -- & -- & -- & 0.1266 \\
& Rubric      & 0.4590 & 0.2670 & 0.5632 & -0.3443 & -- & -- & -- & 0.2362 \\
& Reference   & 0.2972 & 0.2182 & -0.2887 & -0.3288 & -- & -- & -- & -0.0255 \\
& Ref.+Rubric & 0.4790 & 0.2732 & -0.0383 & -0.4047 & -- & -- & -- & 0.0773 \\
\hline

\multirow{4}{*}{DeepSeek-V4-Flash}
& Vanilla     & 0.4007 & 0.2441 & 0.7522 & \textbf{0.8308} & -- & -- & -- & 0.5570 \\
& Rubric      & 0.5110 & 0.3408 & 0.6157 & 0.5908 & -- & -- & -- & 0.5146 \\
& Reference   & 0.4767 & 0.1957 & 0.7362 & 0.6730 & -- & -- & -- & 0.5204 \\
& Ref.+Rubric & 0.5097 & 0.2582 & 0.6426 & 0.4417 & -- & -- & -- & 0.4631 \\
\hline

\multirow{4}{*}{GLM-5.1}
& Vanilla     & 0.4148 & 0.2878 & 0.6635 & 0.7202 & -- & -- & -- & 0.5216 \\
& Rubric      & 0.5392 & 0.3341 & \textbf{0.7733} & 0.5266 & -- & -- & -- & 0.5433 \\
& Reference   & 0.4639 & 0.2552 & 0.7575 & 0.5124 & -- & -- & -- & 0.4973 \\
& Ref.+Rubric & 0.5307 & 0.2772 & 0.5151 & 0.4818 & -- & -- & -- & 0.4512 \\
\hline

\multirow{4}{*}{Kimi-K2.6}
& Vanilla     & 0.4529 & 0.3233 & 0.5163 & 0.7610 & -- & -- & -- & 0.5134 \\
& Rubric      & 0.5018 & 0.3656 & 0.6176 & 0.6084 & -- & -- & -- & 0.5234 \\
& Reference   & 0.4731 & 0.3222 & 0.2185 & 0.3988 & -- & -- & -- & 0.3532 \\
& Ref.+Rubric & 0.4938 & 0.3475 & -0.0502 & 0.3969 & -- & -- & -- & 0.2970 \\
\hline

\end{tabular}%
}
\caption{Spearman's rank correlation coefficient of different LLM-as-a-judge baselines. SurGE reports structure and content quality. Ref.+Rubric denotes reference-based judging with rubric guidance. WP-Bench, Verify, and MA are not included in this correlation analysis, and Avg. averages the available metrics.}
\label{tab:spearman-result}
\end{table*}

\begin{table*}[t]
\centering
\resizebox{\textwidth}{!}{%
\begin{tabular}{llcccccccc}
\hline
\multirow{2}{*}{Model} 
& \multirow{2}{*}{Setting} 
& \multirow{2}{*}{DR-Bench} 
& \multirow{2}{*}{RealDR} 
& \multicolumn{2}{c}{SurGE} 
& \multirow{2}{*}{WP-Bench} 
& \multirow{2}{*}{Verify} 
& \multirow{2}{*}{MA} 
& \multirow{2}{*}{Avg.} \\ 
\cline{5-6}
& & & & Structure & Content & & & & \\ 
\hline

\multirow{4}{*}{\begin{tabular}[c]{@{}l@{}}Qwen3-32B\\(w/o thinking)\end{tabular}} 
& Vanilla     & 0.2366 & 0.2722 & 0.5730 & 0.6588 & -- & -- & -- & 0.4352 \\
& Rubric      & 0.3103 & 0.1707 & 0.6689 & 0.7169 & -- & -- & -- & \textbf{0.4667} \\
& Reference   & 0.3001 & 0.2094 & 0.5798 & \textbf{0.7604} & -- & -- & -- & 0.4624 \\
& Ref.+Rubric & 0.2968 & 0.1970 & 0.5281 & 0.6848 & -- & -- & -- & 0.4267 \\
\hline

\multirow{4}{*}{\begin{tabular}[c]{@{}l@{}}Qwen3-32B\\(w/ thinking)\end{tabular}} 
& Vanilla     & 0.3206 & 0.1843 & 0.2507 & 0.4001 & -- & -- & -- & 0.2889 \\
& Rubric      & 0.3269 & 0.1865 & 0.5542 & 0.3299 & -- & -- & -- & 0.3494 \\
& Reference   & 0.2861 & 0.1986 & 0.5059 & 0.6065 & -- & -- & -- & 0.3993 \\
& Ref.+Rubric & 0.2854 & 0.1616 & 0.4088 & 0.5054 & -- & -- & -- & 0.3403 \\
\hline

\multirow{4}{*}{Qwen3-Max} 
& Vanilla     & 0.3064 & 0.2696 & 0.4305 & 0.5064 & -- & -- & -- & 0.3782 \\
& Rubric      & 0.3627 & 0.2327 & 0.6015 & 0.3700 & -- & -- & -- & 0.3917 \\
& Reference   & 0.3373 & 0.2872 & 0.6413 & 0.5200 & -- & -- & -- & 0.4465 \\
& Ref.+Rubric & 0.3262 & 0.2313 & 0.1696 & 0.5554 & -- & -- & -- & 0.3206 \\
\hline

\multirow{4}{*}{GPT-4o-mini} 
& Vanilla     & 0.2276 & \textbf{0.2916} & -0.1595 & 0.0143 & -- & -- & -- & 0.0935 \\
& Rubric      & 0.2758 & 0.2839 & -0.0001 & -0.1060 & -- & -- & -- & 0.1134 \\
& Reference   & 0.3377 & 0.1963 & -0.0515 & -0.5059 & -- & -- & -- & -0.0058 \\
& Ref.+Rubric & 0.3431 & 0.2279 & 0.1325 & -0.0493 & -- & -- & -- & 0.1636 \\
\hline

\multirow{4}{*}{GPT-5.2}
& Vanilla     & 0.1940 & 0.1427 & 0.0885 & -0.0538 & -- & -- & -- & 0.0929 \\
& Rubric      & 0.3008 & 0.1891 & 0.4762 & -0.2927 & -- & -- & -- & 0.1684 \\
& Reference   & 0.2089 & 0.1552 & -0.2361 & -0.2743 & -- & -- & -- & -0.0366 \\
& Ref.+Rubric & 0.3121 & 0.1937 & -0.0333 & -0.3513 & -- & -- & -- & 0.0303 \\
\hline

\multirow{4}{*}{DeepSeek-V4-Flash}
& Vanilla     & 0.2914 & 0.1866 & 0.6542 & 0.7331 & -- & -- & -- & 0.4663 \\
& Rubric      & 0.3424 & 0.2356 & 0.5355 & 0.4965 & -- & -- & -- & 0.4025 \\
& Reference   & 0.3427 & 0.1532 & 0.6435 & 0.5904 & -- & -- & -- & 0.4325 \\
& Ref.+Rubric & 0.3410 & 0.1766 & 0.5134 & 0.3800 & -- & -- & -- & 0.3528 \\
\hline

\multirow{4}{*}{GLM-5.1}
& Vanilla     & 0.3039 & 0.2114 & 0.5738 & 0.6215 & -- & -- & -- & 0.4277 \\
& Rubric      & \textbf{0.3634} & 0.2297 & \textbf{0.6881} & 0.4502 & -- & -- & -- & 0.4329 \\
& Reference   & 0.3228 & 0.1857 & 0.6668 & 0.4233 & -- & -- & -- & 0.3997 \\
& Ref.+Rubric & 0.3470 & 0.1916 & 0.4346 & 0.4166 & -- & -- & -- & 0.3475 \\
\hline

\multirow{4}{*}{Kimi-K2.6}
& Vanilla     & 0.3227 & 0.2485 & 0.4266 & 0.6466 & -- & -- & -- & 0.4111 \\
& Rubric      & 0.3283 & 0.2612 & 0.5178 & 0.5050 & -- & -- & -- & 0.4031 \\
& Reference   & 0.3398 & 0.2475 & 0.1880 & 0.3199 & -- & -- & -- & 0.2738 \\
& Ref.+Rubric & 0.3249 & 0.2473 & -0.0461 & 0.3160 & -- & -- & -- & 0.2105 \\
\hline

\end{tabular}%
}
\caption{Kendall's rank correlation coefficient of different LLM-as-a-judge baselines. SurGE reports structure and content quality. Ref.+Rubric denotes reference-based judging with rubric guidance. WP-Bench, Verify, and MA are not included in this correlation analysis, and Avg. averages the available metrics.}
\label{tab:kendall-result}
\end{table*}

\section{Guidelines for Expert Annotation}
\label{expert_annotation_guidelines}

To ensure the reliability of \textit{LongJudgeBench}, we establish dataset-specific
annotation and quality-control procedures during benchmark construction.
Since \textit{LongJudgeBench} is constructed from both open-source datasets and our
self-collected data, our guidelines cover two settings: preserving existing
human judgments from prior datasets and constructing new expert judgments for
our self-annotated data.

\noindent\textbf{Preservation of Existing Human Judgments:}
For open-source datasets with existing human annotations, including
DeepResearch Bench, SurGE, WritingPreferenceBench, VerifyBench, and MA, we retain
their original reference judgments and evaluation protocols whenever possible.
We check the consistency of task fields, candidate outputs, reference
information, judgment labels, scoring scales, and metadata during
normalization. Invalid or ambiguous instances, such as tied preference pairs from which a clear reference judgment cannot be derived, are excluded from evaluation.

\noindent\textbf{Construction of New Expert Judgments:}
For self-collected data, we design dataset-specific annotation guidelines. In
RealDeepResearch, two annotators with task-relevant backgrounds independently
score each document on a 0--10 scale along three dimensions: logical
structure, presentation form, and bias checking. Logical structure evaluates
the overall organization, section division, heading-content alignment, and
reasoning coherence. Presentation form evaluates language fluency,
conciseness, readability, and coordination between text and visual elements.
Bias checking evaluates fairness, logical fallacies, one-sidedness,
over-reliance on authority, and potentially misleading statements. The final
score is computed as a weighted sum of the three dimensions. When the two
annotators have a large disagreement, a third annotator is introduced for
adjudication.

\noindent\textbf{Review and Quality Control:}
Before formal annotation, annotators are provided with task-specific rubrics
and examples to ensure a shared understanding of the evaluation criteria.
During annotation, candidate outputs are first evaluated independently by
multiple annotators whenever new human judgments are collected. Disagreements
are resolved through adjudication or discussion according to the dataset
protocol. After annotation and normalization, we further conduct format and
consistency checks to ensure that instructions, candidate outputs, rubrics,
references, metadata, and labels are correctly aligned.

\noindent\textbf{Feedback Mechanism:}
We maintain an iterative feedback mechanism during annotation. Annotators are
encouraged to report unclear criteria, ambiguous cases, or examples that are
difficult to judge. These cases are reviewed and, when necessary, used to
refine the annotation guidelines or clarify the decision rules. This mechanism
helps reduce inconsistency and improves the usability of the guidelines across
different long-form scenarios.

\noindent\textbf{Annotator Compensation:}
All annotators are fairly compensated for their work. For RealDeepResearch,
the total annotation compensation amounted to approximately USD
2,260.

\noindent\textbf{Ethical Considerations:}
We require annotators to follow the provided rubrics and make judgments based
on the quality of the candidate outputs rather than model identity or other
irrelevant factors. We also take measures to reduce potential biases in the
evaluation process, such as anonymizing candidate outputs when applicable and
using independent annotation before adjudication. 
These procedures aim to ensure reliable reference judgments for evaluating LLM-as-a-judge methods across diverse long-form scenarios.
All annotators are informed that their annotations would be used for research and benchmark construction, and they provide consent before participating.

\begin{table*}[]
\centering
\resizebox{\textwidth}{!}{%
\begin{tabular}{l}
\hline
\textbf{The prompt used in the vanilla pointwise setting for DeepResearch Bench.} \\ 
\midrule
\begin{tabular}[c]{@{\ }p{\textwidth}@{\ }}
\textbf{Task Description} \\ 
You are a strict, meticulous, and objective research article evaluation expert. \\
You excel at using specific assessment criteria to thoroughly evaluate research articles, providing precise scores and clear justifications. \\[0.5em]

\textbf{User Prompt Format} \\
Task Background \\
There is a deep research task, and you need to evaluate a research article written for this task. \\

\textless task\textgreater \\
``\{instruction\}'' \\
\textless /task\textgreater \\

Article to Evaluate \\

\textless target\_article\textgreater \\
``\{response\_0\}'' \\
\textless /target\_article\textgreater \\

\textless Instruction\textgreater \\
Your Task \\
Please evaluate the overall quality of the above `\textless target\_article\textgreater' as a response to `\textless task\textgreater'. \\
Please provide an overall score between 0 and 10. \\

Output Format Requirements \\
Please strictly follow the `\textless output\_format\textgreater' below for your evaluation result. \\
Do not include any other unrelated content, introduction, or summary. \\
\textless /Instruction\textgreater \\

\textless output\_format\textgreater \\
\{output\_format\} \\
\textless /output\_format\textgreater \\[0.5em]

\textbf{Output Format} \\
\{\{"overall\_score": 0-10\}\}
\end{tabular} \\ 
\hline
\end{tabular}%
}
\caption{The prompt used in the vanilla pointwise setting for DeepResearch Bench. This setting evaluates a single target article without a rubric or reference information.}
\label{prompt:deepresearch-vanilla}
\end{table*}








\end{document}